# CoLaNET - A Spiking Neural Network with Columnar Layered Architecture for Classification

**Mikhail Kiselev**
Chuvash State University
Cheboxary, Russia
`mkiselev@chuvsu.ru`

March 12, 2025

## Abstract

In the present paper, I describe a spiking neural network (SNN) architecture which, can be used in wide range of supervised learning classification tasks. It is assumed, that all participating signals (the classified object description, correct class label and SNN decision) have spiking nature. The distinctive feature of this architecture is a combination of prototypical network structures corresponding to different classes and significantly distinctive instances of one class (=columns) and functionally differing populations of neurons inside columns (=layers). The other distinctive feature is a novel combination of anti-Hebbian and dopamine-modulated plasticity. The plasticity rules are local and do not use the backpropagation principle. Besides that, as in my previous studies, I was guided by the requirement that the all neuron/plasticity models should be easily implemented on modern neurochips. I illustrate the high performance of my network on a task related to model-based reinforcement learning, namely, evaluation of proximity of an external world state to the target state.

## 1    Introduction and motivation

Despite the numerous approaches to implementation of supervised learning in spiking neural networks (SNN) proposed during last two decades, it still remains hard and actual scientific problem. It can be explained by discrete nature of SNN, the major obstacle for direct application of the error backpropagation principle in the spiking domain. Backpropagation is a very efficient and well-studied algorithm but it requires that the loss function would be differentiable with respect to synaptic weights, that is not true for SNN. Besides that, while SNNs are most efficient when implemented in specialized neuroprocessors like Inltel's Loihi, there are significant problems with implementation of backpropagation on this kind of hardware. At last, SNNs are considered as more biologically plausible models than the traditional neural networks, descendants of perceptrons, but nothing similar to the backpropagation algorithm has been found in the living brain.

For this reason, great attention in the world of SNN is payed to the so-called local learning algorithms. In this approach, the rules for synaptic weight modification are allowed to include only parameters of state and activity of the immediate neighbors of the neuron which the synapse belongs to. The classic example

This research is supported by Russian Science Foundation (grant # 25-21-00126)

of this kind of synaptic plasticity rules is STDP (spike timing dependent plasticity) [13]. However, as it was shown in many works, STDP is most suitable for unsupervised learning [1 - 3]. It is why implementation of unsupervised learning in SNN is more thoroughly studied.

Nevertheless, several approaches to SNN-based supervised learning using only local plasticity rules have been proposed. In this paper, we consider only one sub-class of supervised learning tasks, namely, classification. Respectively, we discuss only SNNs used for classification. A relatively recent review SNN-based algorithms for classification can be found in [4]. More recent review but concentrated on image classification is in [5]. To outline the differences of my learning algorithm from the existing ones, let me explicitly formulate the conditions which I impose on the algorithm. They are consequences of the general requirement of its efficient implementation on neuromorphic hardware interacting with external world in real time:

1. **Only spikes.** All data should have the spiking form. The current presented object is described by spike trains emitted by network's input nodes. The class labels have the form of permanent activity of special input nodes, one node per class, which emit spikes while an object of the respective class is presented. The network makes its decision also in form of activity of its specific neurons, again, – one neuron per class. Thus, the whole classification procedure should be realized inside the SNN. For example, we do not consider the liquid state machine approach, where the network only serves for producing informative features, while the classification itself is made by an external classifier.
2. **Rate/population coding.** We do not consider information coding schemes inconsistent with continuous data flow model. For example, the latency coding is appropriate to code stand-alone independent examples presented to the network (like pictures) but cannot be used in the case of continuous data streams (like videos).
3. **Minimum set of simple operations.** The majority of operations should be computationally cheap (addition, comparison, Boolean operations). Multiplicative operations should be rare.
4. **Only local plasticity rules.** While the majority of existing SNN-based classification algorithms are based on projection of the backpropagation idea to the spiking domain (so-called surrogate gradient methods) it does not permit an efficient implementation on neurochips.
5. **Only local operations.** Only non-blocking operations which can be easily implemented in massively parallel asynchronous platform are allowed. For example, determination of the neuron with highest value of membrane potential in some neuron population does not belong to this class of procedures.

If to apply this filter to the existing SNN-based classification algorithms we will see that almost all of them are sifted out. For example, there exists a class of efficient classification algorithms based on dynamic modification of network topology [6, 7], however it is evident that network restructuring is an expensive operation which can hardly be implemented on neuroprocessors efficiently. The algorithm described in [8] seemingly obeys conditions 1, 2, 4, 5 however it includes many expensive computations. The paper [9] describes a classification algorithm with necessary properties but it was tested on a very small and simple dataset that does not allow making conclusions about its applicability to more realistic tasks.

In this article, I present an SNN-based classification learning algorithm satisfying requirements 1 – 5 and suitable for solution of real world problems. It is achieved due to the special selection of SNN architecture called CoLaNET (Columnar Layered Network) and a novel combination of anti-Hebbian and dopamine-modulated local synaptic plasticity rules. All these features are described in the next two Sections.

## 2    The General Idea of CoLaNET

Firstly, let us discuss the overall network structure (Fig. 1, 2) and general principles of its functioning.

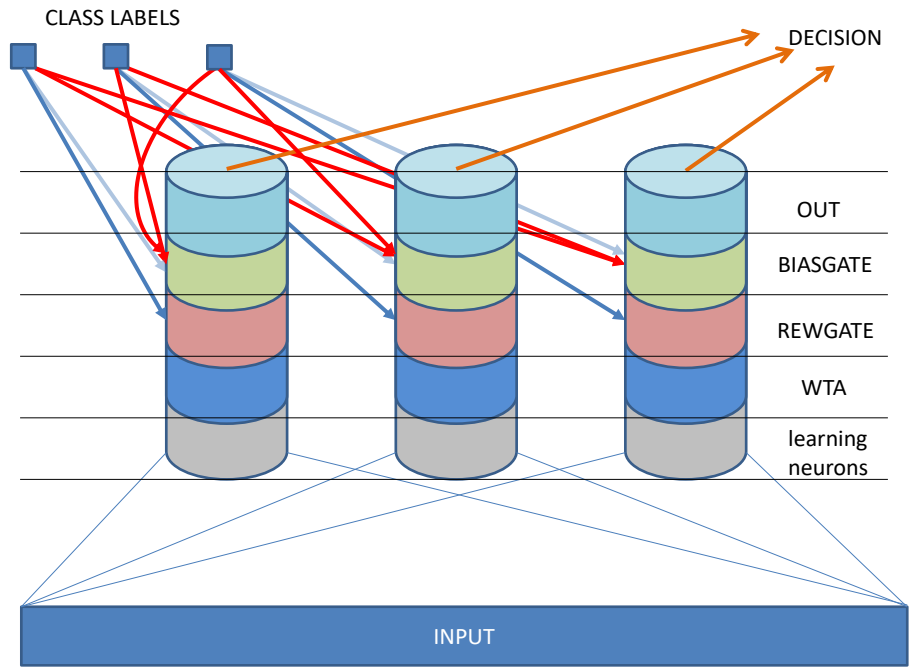

Figure 1. The columnar layered architecture of an SNN implementing classification learning (CoLaNET) – see the detailed description in the text.

The network consists of several identical structures called *columns*. One column corresponds to one target class. Thus, if we apply this SNN to the famous MNIST benchmark there will be 10 columns in the network. Every column contains 5 kinds of neurons organized in 5 *layers* (Fig. 1). The structure of one column is depicted on Fig. 2. We see that it includes several triplets of neurons belonging to the 3 lowest layers. We will call them *microcolumns*. While one column corresponds to one target class, one microcolumn corresponds to significantly distinctive instances (sub-class) of one class. All neurons are described by the simplest LIF (leaky integrate-and-fire) model with slight modifications described later.

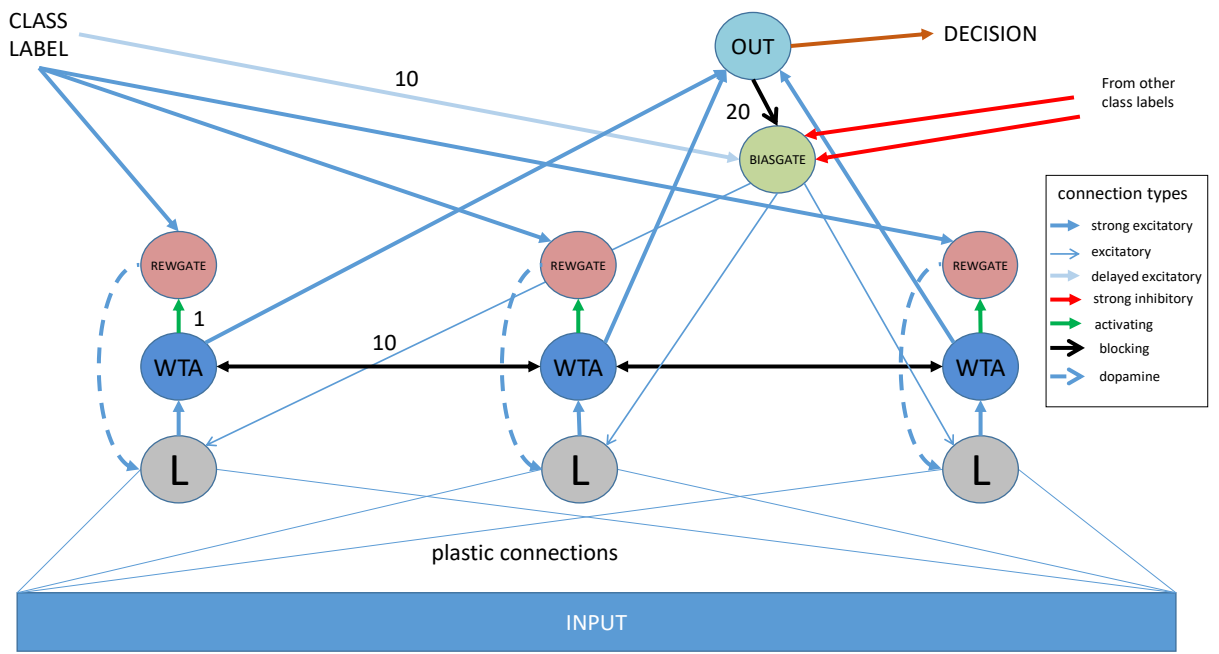

Figure 2. The structure of a single column.

We begin with the inference regime assuming that all neurons have correct values of synaptic weights (which are the result of the learning process considered later). In this regime, the network behavior is simple. The object description has the form of spike trains emitted by input nodes (the blue rectangle at the bottom of Fig. 1 and 2). If the L neurons have right values of their synaptic weights then only an L neuron belonging to the correct column will fire in response to this stimulation. It will cause firing of the WTA neuron in its microcolumn that in its turn will force the correct OUT neuron to fire.

Now discuss the learning process. In this process, the network obtains the information about the current object as well as the information about class (label) of this object. The latter has the form of permanent activity of the respective input node encoding the current class label.

Here, we encounter the first significant distinction from the classification process in traditional neural networks. The traditional formal neurons have no internal dynamics – their output depends only on the current values of their inputs and does not depend on the previous input values. In contrast, spiking neurons are dynamical systems – their state depends on their history. Therefore, in case when the consecutive examples presented to the network are independent (as they are in the typical machine learning tasks), the presentations of two consecutive examples should be separated by a certain period of "silence" – absence of any spikes. It is necessary to exclude influence of the previous object on classification of the current object. In our case, every example is presented during 10 SNN simulation steps (we will assume that 1 simulation step = 1 msec), and the silence period is also 10 msec. It is important that a class label node emits spikes during the whole 20 msec period including the object presentation and the silence.

At the beginning of the learning process, all weights of all plastic synapses (only neurons from the lowest layer have plastic synapses) are zero. Therefore, the stimulation from the input cannot make them fire. However, the SNN has another source of spikes – one of the class label nodes. It sends spikes to the REWGATE neurons of its column. However these neurons are in the inactive state (see the discussion below about the active/inactive neuron states) and are not able to fire. Besides that, this node sends spikes to strong excitatory synapse of the BIASGATE neuron in its column. The connection between it and this neuron is slow – the spikes pass it for 10 msec. Therefore, the first spike from the train emitted by the class label nodes reaches BIASGATE when the input stimulation has ended. This spike train induces constant firing of BIASGATE. The BIASGATE neuron is connected with the learning neurons by the excitatory synapse with the weight sufficient to force neuron to fire just before the end of the silence period (by the constant stimulation from BIASGATE). Since at the beginning, all learning neurons are identical all they fire simultaneously (only in the stimulated column, of course). All they send powerful stimulation to the WTA neurons. But the WTA layer in one column (WTA means "winner takes all") is designed so that no more than one neuron can fire simultaneously. Some random WTA neuron fires (it makes the simulation non-deterministic!). It activates the REWGATE neuron in its microcolumn. But the REWGATE neurons still obtain stimulation from the class label node. Therefore, one REWGATE neuron immediately fires. It emits so called dopamine spike coming at the special dopamine synapse of the learning neuron in the same microcolumn. This spike triggers the *dopamine plasticity* process. The dopamine plasticity rule says that all plastic synapses having obtained spikes some time before neuron firing are potentiated if the neuron receives a dopamine spike shortly after that firing. As a result of this process, one learning neuron in the column corresponding to the class presented gets slightly potentiated synapses connected to the recently active input nodes. These weights are still insufficient for firing solely from the input stimulation. However, the next time when similar stimulus will be presented, this winning neuron will have positive value of the membrane potential at the beginning of stimulation from the BIASGATE. Therefore, it will have high chances to become a winner again, thus further potentiating the same set of synapses.

Here, I should emphasize on another important feature of CoLaNET – different L neurons in one column should recognize significantly different instances of the target class. It is necessary to cover the whole set of target class objects with sufficient accuracy. This goal can be reached by two different ways, they both are used in CoLaNET. The first way is introduction of competition between plastic synapses due to constancy of the total synaptic weight of one neuron. Whenever some synapses are strengthened, all the other synapses are uniformly weakened. It means that the neuron – winner not only becomes more sensitive to the first presented stimulus but also becomes less sensitive to significantly different stimuli. The second way is threshold potential variation. In this approach, the neuron threshold potential is not constant but is proportional to the sum of positive weights of the neuron's plastic synapses. It makes more trained neurons less excitable – they need that stimulation would be more exactly projected to a limited set of strong synapses corresponding to the sub-class recognized. Due to these mechanisms, if the second presented object from the same class will have little resemblance with the first one, then the first winner will have the starting value of the membrane potential (before its forced stimulation from BIASGATE) farther from the threshold potential than the other L neurons in the same column, and, therefore, the WTA neuron in its microcolumn will not win this time. In such a way, L neurons in different microcolumns learn to react to different instances of a target class.

After some number of the plasticity acts described above, some L neurons acquire the ability to fire in response to input stimulation without help of BIASGATE neurons. In this case, the WTA neuron connected to the firing L neuron stimulates the OUT neuron of this column. It fires and blocks the BIASGATE neuron for all period of current object presentation (including the silence period) because stimulation from BIASGATE is not needed now.

It remains to say that this scheme also has protection against wrong L neuron firing. In fact, my plasticity model consists of two components – anti-Hebbian plasticity and dopamine plasticity. Dopamine plasticity was briefly described above. The anti-Hebbian plasticity mechanism is also simple. Whenever the neuron fires, all its plastic synapses having received a spike shortly before this are depressed. It is just the contrary to the original Hebbian law stating that all synapses helping the neuron to fire are potentiated. But in our case, the anti-Hebbian rule is needed. Indeed, L neurons should react only to the correct stimuli. The correct stimuli are marked by the activity of the respective class label node which causes dopamine reward of the L neuron. If an L neuron fired and did not receive the dopamine reward, it fired wrongly and, therefore, the synapses which forced it to fire should be suppressed. Thus, the complete picture is the following. When an L neuron fires (and this firing is not forced by a strong non-plastic synapse) all its synapses which contributed to this firing are depressed. They remain depressed if nothing more happens. But if, afterwards, this neuron receives a dopamine spike these synapses are potentiated. Hence, three possible scenarios are possible:

1. **The neuron did not fire during input stimulation and was selected as a target for stimulation from BIASGATE.** Only dopamine plasticity should work – to potentiate the synapses receiving spikes. It gives it chances to fire correctly next time.
2. **The neuron fired during input stimulation but it was wrong (no dopamine reward).** Only anti-Hebbian plasticity works - synapses receiving spikes are depressed. It lowers the neuron's chances to fire wrongly next time.
3. **The neuron fired during input stimulation but it was right (dopamine reward followed).** Both plasticity mechanisms work – but they work in the opposite directions so that nothing changes. The neuron works correctly – we should not change it.

This informal description shows schematically how this SNN learns. In the next Section, I will describe this process in more formal way.

## 3    Models of Neuron and Plasticity

Let us consider the models of neuron and synaptic plasticity more formally. In this research, I use LIF (leaky integrate-and-fire) neuron model. It is a very simple and frequently used neuron model. Besides that, it is efficiently implementable on the modern neurochips (such as TrueNorth, Loihi, AltAI).

### 3.1 The Simplest LIF Neuron

The simplest current-based delta synapse model is used for all excitatory and inhibitory synapses. It means that every time the synapse receives a spike, it instantly changes the membrane potential by the value of the synaptic weight (positive or negative – depending on the synapse type). Thus, the state of a neuron at the moment $t$ is described by its membrane potential $u(t)$ whose dynamics are defined by the equation

$$\frac{du}{dt} = -\frac{u}{\tau_v} + \sum_{i,j} w_i \delta(t - t_{ij}) \tag{1}$$

and the condition that if $u$ exceeds the threshold potential $h$ then the neuron fires and value of $u$ is decremented by $h$. The meaning of the other symbols in (1) is the following: $\tau_v$ – the membrane leakage time constant; $w_i$ - the weight of $i$-th synapse; $t_{ij}$ - the time moment when $i$-th synapse received $j$-th spike.

The synapses which can change the membrane potential may be plastic or fixed. Setting the correct values of plastic synapses is the aim of the learning process. Fixed synapses are usually strong and serve for the correct organization of the learning process. Neuron firing caused by a spike coming at one of its fixed synapses will be called *forced* firing (see the discussion of anti-Hebbian plasticity below).

### 3.2 Gating Synapses and Neuron Inactivation

As it was said above, I introduced only one non-standard feature in the LIF model (for the neurons WTA, REWGATE and BIASGATE) – the active/inactive neuron functioning regimes. In the active regime, a neuron behaves like a normal LIF neuron obeying (1). In the inactive regime, presynaptic spikes do not change value of the membrane potential. The current neuron regime is determined by the sign of the neuron state component called the *activity time a*. The neuron is active if $a > 0$. Every simulation tact, $a$ changes in accordance with the following formula:

$$a \leftarrow \begin{cases} a + 1 \; if \; a < -1 \\ +\infty \; if \; a = -1 \\ 0 \; if \; a = 0 \\ a - 1 \; if \; a > 0 \end{cases}. \tag{2}$$

At the simulation beginning, activity time of all neurons except REWGATE neurons is set to a very high positive value ($+\infty$) and, therefore, these neurons are active. For REWGATE neurons, $a$ is set to 0. Therefore, they are not active. Neurons may have special gating synapses. A spikes coming to a gating synapse with the weight $\omega$ changes $a$ by the following rule:

$$a \leftarrow \begin{cases} \min(a, \omega) \; if \; \omega < 0 \\ \max(a, \omega) \; if \; \omega > 0 \end{cases}. \tag{3}$$

It can be easily seen that this logic of $a$ modification leads to the scheme of CoLaNET neuron interactions described in the previous section. To check it, let us formulate specifications on the behavior of neurons in CoLaNET and show how they are satisfied by the temporal parameters of neurons and connections (they are denoted by the numbers on Fig. 2).

1. **At most one WTA neuron should fire in one column during one object presentation (including the silence period).**
2. **At most one dopamine plasticity act should happen in one column during one object presentation.** This requirement is satisfied if p.1 is true and the weight of the activating connection between WTA and REWGATE equals to 1.
3. P.1 is true if the weight of the connections between WTA neurons equals to 10. Indeed, if an L neuron fires during object presentation then it is guaranteed that only one WTA neuron will fire in the object presentation period (since its length is 10 msec). An L neuron might be forced to fire by stimulation from BIASGATE in the silence period however BIASGATE is blocked by WTA neuron firing (via the OUT neuron) for 20 msec. Strictly speaking, such a long inactivity period of BIASGATE may lead to undesirable interference between consecutive presentations of objects belonging to the same class. Indeed, if some L neuron is forced to fire by BIASGATE during the silence period, then this BIASGATE neuron will be blocked for 20 msec. This time interval partially overlaps the silence period of the next object presentation (if the next objects is from the same class). It will lead to lower stimulation level from BIASGATE which may become insufficient for forced L firing – in this case, this object presentation will not be used for learning. This negative effect could be remedied by extension of silence period to, say, 20 msec, but it would make the learning process 1.5 slower. So that, I consider it wiser to retain all the timings as they are.

Also it is obvious, that this gating synapse mechanism can be easily implemented in hardware as it requires only increment/decrement and comparison operations.

## 3.3 Synaptic Resource

Now, let us consider the synaptic plasticity model used in CoLaNET. Its main distinctive feature is the same as in our previous research works [10 - 12]. Namely, synaptic plasticity rules are applied to the so called *synaptic resource W* instead of the synaptic weight *w*. There is functional dependence between *W* and *w* expressed by the formula

$$w = w_{\min} + \frac{(w_{\max} - w_{\min})\max(W,0)}{w_{\max} - w_{\min} + \max(W,0)}, \qquad (4)$$

where $w_{min}$ and $w_{max}$ are constant.

In this model, the weight values lay inside the range [$w_{min}$, $w_{max}$] - while *W* runs from -∞ to +∞, *w* runs from $w_{\min}$ to $w_{\max}$. The arguments in favor of this approach are discussed in [10 – 12].

As it was said in Section 2, the CoLaNET synaptic plasticity model comprises two distinct and independent components – anti-Hebbian plasticity and dopamine plasticity. They are considered in two next subsections.

## 3.4 Anti-Hebbian Plasticity

The standard STDP (spike timing dependent plasticity) model [13] states that spikes coming short time before postsynaptic spike emission potentiate the synapses receiving them. This concept aligns with Donald Hebb's principle, which asserts that synaptic plasticity should reflect causal relationships between neuron firings; so that the synapses inducing neuron firing should be strengthened. Plenty of neurophysiological observations have proved this principle. However, in-depth investigations into plasticity within biological neurons have revealed multiple instances of entirely distinct synaptic plasticity models existing in nature [14, 15]. Furthermore, examples of plasticity rules acting in the direction opposite to Hebbian principle (anti-Hebbian plasticity) have been observed in different organisms [16]. It makes us conclude that different kinds of synaptic plasticity are suitable for the solution of different problems.

As it was discussed in Section 2, in our case, the anti-Hebbian plasticity model is more appropriate. It is described by the following very simple rule. Resources of all synapses having obtained at least one spike during the time $T_H$ before the neuron firing are decreased by the constant value $d_H$ if this firing is not forced.

## 3.5 Dopamine Plasticity

As it is shown on Fig. 2, every L neuron has a plasticity-inducing (dopamine) synapse connecting the L neuron with the GATEREW neuron in the same microcolumn. The dopamine plasticity mechanism is applied to a neuron when it receives a spike via its dopamine spike but only if this neuron fired not earlier than the time $T_P$ before this. In this case, resources of all synapses having obtained at least one spike during the time $T_H$ before that neuron firing are increased by the constant value $d_D$. For dopamine plasticity, it does not matter was this firing forced or not.

It was noted in Section 2 (see point 3) that the neuron which fires correctly should be considered as already learnt and, therefore, should not change anymore. This condition is satisfied if $d_D = d_H$. But in the case of many noisy training examples, it found to be useful if $d_D$ is somewhat greater than $d_H$.

### 3.6 Constant Total Synaptic Resource

As it was said, in order to introduce competition between synapses inside one neuron and between neurons inside one WTA group, I added one more component to the model of synaptic plasticity – constancy of neuron's total synaptic resource. Whenever some synapses are depressed or potentiated due to the above mentioned plasticity rules all the other synapses are changed in the opposite direction by the constant value equal for all these synapses such that the total synaptic resource of the neuron is preserved. Effect of this rule can be controlled introducing imaginary unconnected synapses whose only role is to be a reservoir for the excessive (or additional) resource. The competitive effect is maximum when there are no such silent synapses and it vanishes with their number approaching infinity.

At this point, I described the model of neuron and synaptic plasticity completely and now we will consider one example of classification problems solved by CoLaNET.

### 3.7 Dependency of Threshold Potential on Total Positive Synaptic Weight

It is another method to implement competitive learning of L neurons inside one column. In my model, the threshold potential $h$ is a sum of the base value equal to 1 and the variable part proportional to the sum of positive synaptic weights of the neuron:

$$h = 1 + \alpha \sum_i \max(0, w_i), \tag{5}$$

where $\alpha < 1$ is a small constant.

## 4 The Test Classification Task – the Target States in the Ping-Pong ATARI Game.

One of possible applications of CoLaNET – evaluation of the current state of the external world in model-based reinforcement learning (RL). It is necessary to determine how close the agent to its goal. As an example, I took one of the well-known RL benchmarks – the ping-pong ATARI game [17]. In this game, a ball traverses within a square area, rebounding off its walls. The area has only three walls. Instead of the left wall, the racket moves in the vertical direction on the left border of this square area. The racket is controlled by the agent, which can move it up and down. When the ball hits the racket, it bounces back and the agent obtains a reward signal. If the ball crosses the left border without hitting the racket the agent gets punishment and the ball is returned to a random point of the middle vertical line of the area, gets random movement direction and speed and the game continues. Using the reward/punishment signals received, the agent should understand that its aim is to reflect the ball and learn how to do it.

In our example, the network's task is distinguishing the states when it is not needed to move the racket because the ball will bounce from it from the states when racket movements are necessary.

Let us consider the input information coming to plastic synapses of the L neurons. This information includes the current positions of the ball and the racket and the ball velocity. While the ultimate formulation of this problem would involve primary raster information (i.e., the screen image), computer vision is not in our primary focus in this study. Consequently, we assume that preceding network layers have already processed the primary raster data and converted it into the spike-based description of the world state, which forms the basis of the L neuron input.

The input nodes that are sources of spikes sent to the learning neurons are subdivided into the following sections:

- The ball X coordinate Consists of <u>30 nodes</u> capturing the ball's horizontal position. The horizontal dimension is broken to 30 bins. When the ball is in the bin $i$, the $i$-th node emits spikes with frequency 300 Hz. To establish spatial and temporal scales we assume that the size of the square area is 10×10 cm (so that the boundary coordinates are ±5 cm) and the discrete emulation time step is 1 msec.
- The ball Y coordinate. Consists of <u>30 nodes</u> capturing the ball's vertical position. Similar to X but for the vertical axis.
- The ball velocity X component. Consists of <u>9 nodes</u> capturing the ball's horizontal velocity. When the ball is reset in the middle of the square area, its velocity is set to the random value from the range [10, 33.3] cm/sec. Its original movement direction is also random but it is selected so that its X component would not be less than 10 cm/sec. The whole range of possible ball velocity X component values is broken to 9 bins such that the probabilities to find the ball at a random moment in each of these bins are approximately equal. While the ball X velocity is in some bin, the respective input node emits spikes with 300 Hz frequency.
- The ball velocity Y component. Consists of <u>9 nodes</u> capturing the ball's vertical velocity. The same logic as for the X velocity component.
- The racket Y coordinate. Consists of <u>30 nodes</u> capturing the racket's vertical position. Similar to the ball Y coordinate. The racket size is 1.8 cm so that the racket takes slightly more than 5 vertical bins.
- The relative position of the ball and the racket in the close zone. Consists of <u>25 nodes</u> capturing the ball's positions close to the racket. The square visual field of size 3×3 cm moves together with the racket so that the racket center is always at the center of the left border of this visual field. The visual field is broken to 5×5 square zones. When the ball is in some zone, the respective input node fires with frequency 300 Hz.

In total, there are 133 input nodes transmitting their spikes to the learning neurons. The SNN's objective was to discern the world states which lead to obtaining reward during next $T$ msec without racket moving. After the ball hits the left wall or the racket, the ball and racket positions are randomly reset. This is a binary classification problem therefore only one column is needed here. Examples of non-binary classification tasks (such as MNIST) are considered in a separate paper [20].

In order to apply a CoLaNET network to this task, the input spiking data were recorded in a file. They were preprocessed in the following way. The whole ping-pong simulation took 2000 sec. Only those its fragments were left which were not earlier that $T$ msec before the next ball hit to the left wall (or the racket). These fragments were cut to 10 msec intervals. They were randomly shuffled. 10 msec intervals of silence were inserted between them (see the discussion in Section 2). Of course, the correct labels (good/bad state) were retained for each interval. 2/3 of these data were used for training the network. After that, all weights got fixed and the network accuracy was tested on the rest of the data. The F measure was selected as an accuracy criterion. The example of SNN with CoLaNET structure written in ArNI-X language [18] can be found in Appendix A.

The CoLaNET architecture has a very few hyperparameters (see Table 1). In order to find their optimum values I used an optimization procedure based on genetic algorithm. To diminish the probability of accidental bad or good result, I averaged the criterion (the absolute error value) for 4 tests with the same SNN parameters. Since the WTA mechanism introduces non-determinism in the simulation process, the synaptic weights learnt in these tests were different. I used the typical genetic algorithm settings, used by me in my previous studies. The population size was 100, mutation probability per individual equaled to 0.5, elitism level was 0.1. Genetic algorithm terminated after 3 consecutive populations without optimization progress.

The parameters varied in the optimization procedure, the ranges of their variation and their optimum values are presented in Table 1. The problem was solved for $T = 300$ msec.

**Table 1. The CoLaNET hyperparameters optimized.**

| Parameter optimized | Value range | Optimum value |
|---|---|---|
| Learning rate $d_D$ | 0.004 – 0.4 (everywhere in this table the value of the threshold membrane potential is taken equal to 1) | 0.0186 |
| Ratio $d_H / d_D$ | 0 – 1 | 0.582 |
| Maximum input weight $w_{max}$ | 0.04 – 0.4 | 0.328 |
| Maximum input weight $w_{min}$ | -0.0004 – -0.4 | -0.00746 |
| Number of microcolumns in one column | 1 – 30 | 2 |
| Threshold potential variability $\alpha$ | 0.001 – 0.3 | 0.005525 |

The result was reached at the 12[th] generation. The winning network showed F measure equal to 0.452 ± 0.03. Interestingly, the precision was significantly higher than the recall (0.66 against 0.34).

The weights learnt in this task are depicted on Figure 3. The learning results are presented on Fig. 5 depicting values of synaptic resources of the L neurons at the end of the learning period. The leftmost plots correspond to 30 input nodes coding the ball X coordinate. The vertical axis of all plots except the rightmost ones displays synaptic resource value. The second plot column corresponds to 30 input nodes coding the Y coordinate of the ball (the blue line) and the racket (the orange line). The next two plot columns represent 9+9 input nodes coding the horizontal and vertical components of the ball velocity.

The rightmost plot column shows the color-coded values of synaptic resources of the 25 input nodes indicating the location of the ball within a 5x5 grid that moves with the racket. The distribution of synaptic resource values in these plots appears reasonable and in line with expectations. We see different examples of "good" states. For example, the topmost L neuron corresponds to the case when the ball moves almost horizontally to left and the racket Y coordinate is close to the ball Y coordinate. The 2[nd] neuron fires when the ball is higher than the racket, is near the left wall and has the velocity vector directed left and down.

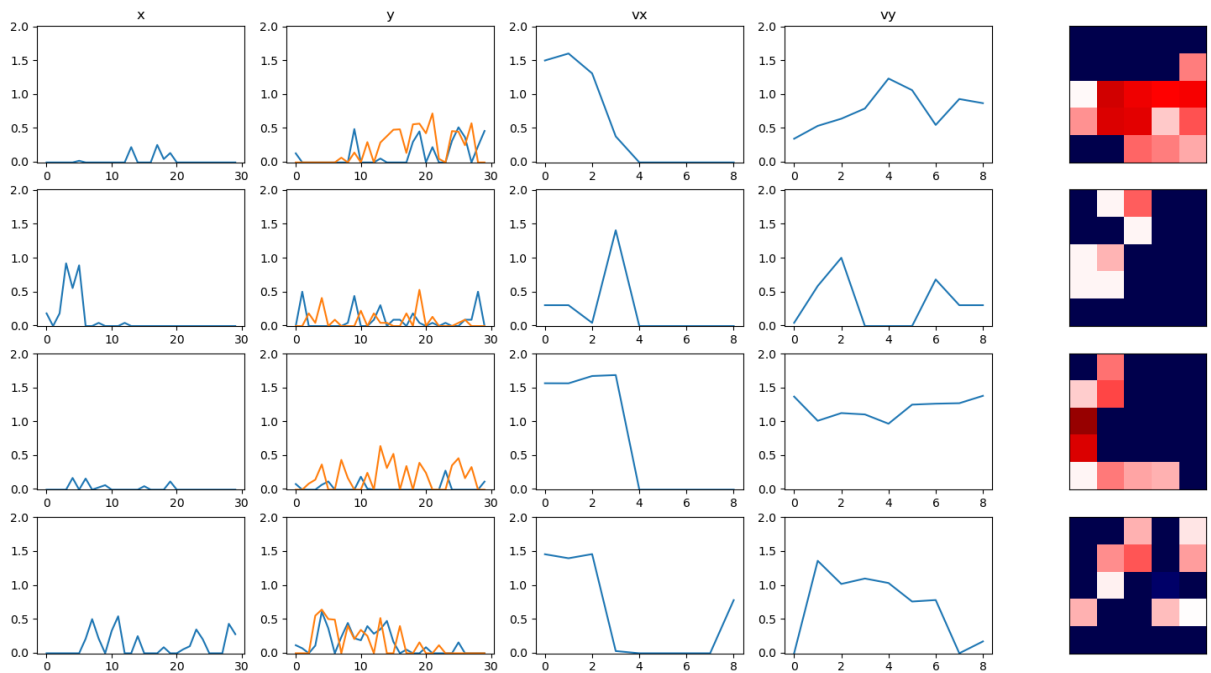

Figure 3. The learnt weights of neurons in the best SNN at the learning period end.

To objectively evaluate the performance of CoLaNET on this task, I compared it with one of the most accurate traditional machine learning algorithms, namely, random forest (from the Sklearn package [19]). Random forest was trained on the same binary signal data from the input nodes, with each 10 msec interval serving as a learning example with the same division to learning and testing sets. Random forest showed F measure only slightly above the CoLaNET result – 0.48, with the same ratio of precision/recall (0.71/0.36). However, it is important to stress that random forest model was obviously bigger in terms of degrees of freedom. It contained 100 trees, while CoLaNET total weight count was 4 * 133 = 532. Another method used by me to assess quality of CoLaNET solution was determination of theoretical limit for F measure in this task. Indeed, if we know the current discrete state of the whole system (from the current activity of the input nodes) then we can evaluate the mean expected values of all coordinates and velocities and thus decide whether the current state is good or bad. Naturally, this decision will not be exact since we do not know exact coordinates and velocities, but it will help determine the upper limit of accuracy in this task. This procedure gives 0.56 precision and 0.65 recall, and, correspondingly, F measure equal to 0.6. I think that these results can be considered as an evidence in favor of CoLaNET efficiency. In further works, we will test CoLaNET on the tasks from computer vision.

# 5    Conclusion.

In this paper, I describe a novel SNN architecture called CoLaNET for solution of classification problems. It is made in purely spiking fashion – all data has the spiking form and all processing is performed by the SNN. It is especially important from the view point of application of modern neurochips where all data are represented in the form of spikes transmitted as AER packets. In general, efficient implementation of synapse/neuron/network models on this kind of neurochips was keynote of this study.

CoLaNET has a specific structure combining columns and microcolumns and 5 layers of neurons such that neurons on each layer have similar properties and connectivity patterns. I do not hypothesize that this structure resembles to some extent the columnar layered organization of neurons in the neocortex but I do not exclude this idea – it requires further checks.

CoLaNET consists of LIF neurons with a simple addition in the form of gating synapses. The learning process is based on a novel combination of anti-Hebbian and dopamine plasticity. To evaluate efficiency of CoLaNET, I applied it to a problem from the field of reinforcement learning – evaluation of the current racket position in the ping-pong ATARI game. The accuracy demonstrated by CoLaNET was found to be close to random forest, one of the best modern ML algorithms, and not far from the upper theoretical limit.

Further development and exploration of the CoLaNET architecture is planned in the future in the following directions:

- Modification of CoLaNET for solution of regression problems.
- Incorporation of CoLaNET in the SNN implementing model-based RL.
- Application of CoLaNET to wide range of classification tasks including the practical ones.
- Directed inference of CoLaNET hyperparameters from dataset characteristics (without expensive optimization procedures).
- Exploration of CoLaNET from viewpoint of the catastrophic forgetting problem and stability with respect to noise.
- Rigor mathematic exploration of CoLaNET properties.
- Combination of CoLaNET with convolutional SNNs.
- Hierarchical extension of CoLaNET.
- Exploration of possible implementation of CoLaNET on the modern neuroprocessors.

## Acknowledgements.

The present work is a part of the research project in the field of SNN carried out by Chuvash State University. Access to Loihi-based computational systems is provided as a part of Chuvash State University participation in Intel's INRC program. The idea of columnar SNN organization appeared in the project Kaspersky Neuromorphic Platform (KNP) in which I work as a member of Kaspersky Neuromorphic AI team. Further, the work on columnar SNNs was continued by me in the RL-related research projects conducted by me at Cifrum company, the Laboratory of Neuromorophic Computations of Chuvash State University and Kaspersky. This work and discussions with Andrey Laverntyev, Denis Larionov, Dmitry Ivanov and Vladimir Klinshov were very fruitful for formation of the present variant of CoLaNET.

My SNN simulator package ArNI-X was used to obtain all results reported in this paper.

## References.

# Appendix A. An Example of CoLaNET Configuration on the ArNI-X Language

```xml
<?xml version="1.0" encoding="utf - 8"?>
<SNN>
  <Global>0</Global>
  <Global>0.00552501</Global>
  <RECEPTORS name="R" n="133">
    <Implementation lib="fromFile">
      <args type="text">
        <source>inpstaticperm.txt</source>
      </args>
    </Implementation>
  </RECEPTORS>
  <RECEPTORS name="Target" n="1">
    <Implementation lib="StateClassifier">
      <args>
        <target_file>inpstatictargetperm.txt</target_file>
        <spike_period>1</spike_period>
        <state_duration>20</state_duration>
        <learning_time>748940</learning_time>
    <no_class>0</no_class>
    <criterion>averaged_F</criterion>
    <sequential_test></sequential_test>
    <prediction_file>tmp.csv</prediction_file>
      </args>
    </Implementation>
  </RECEPTORS>
<NETWORK ncopies="1">
    <Sections>
        <Section name="L">
          <props>
        <n>4</n>
        <Structure type="O" dimension="1"></Structure>
        <chartime>3</chartime>
        <weight_inc>-0.109361</weight_inc>
        <dopamine_plasticity_time>10</dopamine_plasticity_time>
        <maxTSSISI>10</maxTSSISI>
        <stability_resource_change_ratio>1.30805</stability_resource_change_ratio>
        <minweight>-0.0108558</minweight>
        <maxweight>2.00498</maxweight>
        <three_factor_plasticity></three_factor_plasticity>
        <nsilentsynapses>10</nsilentsynapses>
        <hebbian_plasticity_chartime_ratio>3.59994</hebbian_plasticity_chartime_ratio>
          </props>
        </Section>
        <Section name="WTA">
          <props>
        <n>4</n>
        <Structure type="O" dimension="1"></Structure>
        <chartime>1</chartime>
          </props>
        </Section>
        <Section name="REWGATE">
          <props>
        <n>4</n>
        <Structure type="O" dimension="1"></Structure>
        <chartime>1</chartime>
          </props>
        </Section>
        <Section name="OUT">
          <props>
        <n>1</n>
        <chartime>1</chartime>
          </props>
        </Section>
        <Section name="BIASGATE">
          <props>
        <n>1</n>
```

```xml
        <chartime>1</chartime>
          </props>
        </Section>
        <Link from="R" to="L" type="plastic">
          <IniResource type="uni">
        <min>0.011</min>
        <max>0.011</max>
          </IniResource>
            <probability>1</probability>
          <maxnpre>1000</maxnpre>
        </Link>
        <Link from="L" to="WTA" policy="aligned">
          <weight>9</weight>
        </Link>
        <Link from="WTA" to="WTA" policy="all-to-all-sections" type="gating">
          <weight>-10</weight>
        </Link>
        <Link from="WTA" to="REWGATE" policy="aligned" type="gating">
          <weight>1</weight>
        </Link>
        <Link from="REWGATE" to="L" policy="aligned" type="reward">
          <weight>0.158111</weight>
        </Link>
        <Link from="WTA" to="OUT" policy="aligned">
          <weight>10</weight>
        </Link>
        <Link from="OUT" to="BIASGATE" policy="aligned" type="gating">
          <weight>-10</weight>
        </Link>
        <Link from="Target" to="REWGATE" policy="aligned">
          <weight>10</weight>
        </Link>
        <Link from="Target" to="BIASGATE" policy="aligned">
          <weight>10</weight>
          <Delay type="uni">
        <min>10</min>
        <max>10</max>
          </Delay>
        </Link>
        <Link from="Target" to="BIASGATE" policy="exclusive">
          <weight>-30</weight>
        </Link>
        <Link from="BIASGATE" to="L" policy="aligned">
          <weight>3</weight>
        </Link>
      </Sections>
  </NETWORK>
    <Readout lib="StateClassifier">
      <output>OUT</output>
    </Readout>
    </SNN>
```